\def\year{2018}\relax
\documentclass[letterpaper]{article} 
\usepackage{aaai18}  
\usepackage{times}  
\usepackage{helvet}  
\usepackage{courier}  
\usepackage{url}  
\usepackage{graphicx}  
\usepackage{booktabs}
\usepackage{amsmath}
\usepackage{amssymb}
\usepackage{multirow}
\usepackage{color}
\usepackage{soul}

\newcommand\blfootnote[1]{%
  \begingroup
  \renewcommand\thefootnote{}\footnote{#1}%
  \addtocounter{footnote}{-1}%
  \endgroup
}

\def\figref#1{Fig.~\ref{#1}}

\def\tabref#1{Table~\ref{#1}}
\def\eqnref#1{Eqn.~\ref{#1}}

\newcommand\subjobjswap[0]{\mbox{\textsc{SOSwap}}}
\newcommand\amodswap[0]{\mbox{\textsc{AddAmod}}}

\newcommand{\citet}[1]
{\citeauthor{#1} ̃\shortcite{#1}}
\newcommand{\citep}{\cite}

\pdfoutput=1

\begin{document}
\title{Analyzing Compositionality-Sensitivity of NLI Models}
\author{Yixin Nie\textsuperscript{*}\;\;\;\; Yicheng Wang\textsuperscript{*}\;\;\;\; Mohit Bansal\\
Department of Computer Science\\
University of North Carolina at Chapel Hill\\
\texttt{\{yixin1, yicheng, mbansal\}@cs.unc.edu}\\
}
\maketitle

\blfootnote{\textsuperscript{*} indicates equal contribution.}

\begin{abstract}
Success in natural language inference (NLI) should require a model to understand both lexical and compositional semantics.
However, through adversarial evaluation, we find that several state-of-the-art models with diverse architectures are over-relying on the former and fail to use the latter. 
Further, this compositionality unawareness is not reflected via standard evaluation on current datasets.
We show that removing RNNs in existing models or shuffling input words during training does not induce large performance loss despite the explicit removal of compositional information.
Therefore, we propose a compositionality-sensitivity testing setup that analyzes models on natural examples from existing datasets that cannot be solved via lexical features alone (i.e., on which a bag-of-words model gives a high probability to one wrong label), hence revealing the models' actual compositionality awareness. 
We show that this setup not only highlights the limited compositional ability of current NLI models, but also differentiates model performance based on design, e.g., separating shallow bag-of-words models from deeper, linguistically-grounded tree-based models.
Our evaluation setup is an important analysis tool: complementing currently existing adversarial and linguistically driven diagnostic evaluations, and exposing opportunities for future work on evaluating models' compositional understanding.

\end{abstract}

\section{Introduction}
Natural Language Inference (NLI)
is a task in which a system is asked to classify the relationship between a pair of premise and hypothesis as one of either entailment, contradiction, or neutral. 
This task is considered to be the basis of many downstream, higher-level NLP tasks that require complex natural language understanding such as question-answering and summarization.
Large annotated datasets such as the Stanford Natural Language Inference \cite{snli:emnlp2015} (SNLI) and the Multi-Genre Natural Language Inference \cite{williams2017broad} (MNLI) have promoted the development of many different neural NLI models, including encoding and co-attention models using both sequential and recursive representations~\cite{RSE,choi2017learning,ESIM,DIIN,DRLSTM,decomposable,bimpm}, all achieving near human-level performance on standard datasets.

Despite their high performance, it is unclear if these models employ semantic understanding of natural language to classify these pairs, or are simply performing word/phrase-level pattern matching.
We first conduct investigative experiments with rule-based adversaries and show empirically that seven state-of-the-art NLI models, spanning a variety of architectures, are all unable to recognize simple semantic differences when the word-level information remains unchanged (e.g., the swapping of the subject and object or the addition of the same modifier to different governors).\footnote{We release our adversaries, compositionality-sensitivity testing setups, and all code at \scriptsize\url{https://github.com/easonnie/analyze-compositionality-sensitivity-NLI}.} Their failure on these examples contrasts sharply with their high performance on the standard evaluation set, indicating that standard evaluation does not sufficiently assess sentence-level understanding.

Next, to further show the insufficiency of standard evaluation for testing compositional understanding, we conduct two additional experiments in which the compositional information is removed or diluted. Firstly, we train and evaluate the state-of-the-art models with their RNNs replaced by fully-connected layers. Secondly, we train these models with input words shuffled and evaluate them on the original evaluation datasets. 
In both of the experiments, models are still able to achieve high performance on the standard evaluation datasets, demonstrating that standard evaluation is unable to sufficiently separate models with compositionality understanding capabilities from those without.

We also show the limitation of adversarial evaluation setups by demonstrating their narrow scope:
each type of adversary is only capable of testing a model's ability to process one specific type of compositional semantics.
We show via adversarial training that success on one type of adversary does not generalize to other types of adversaries, but instead induces errors caused by over-fitting the training data.
Thus, while adversarial evaluation is useful for exposing the issue of lexical over-stability, it is not a robust measure of models' ability to understand semantic compositional information.

Hence, in order to analyze a model's abilities to reason beyond the lexical level and reveal its sensitivity to compositional differences, we present a `compositionality-sensitivity' testing setup: we select examples for which a bag-of-words model is misguided (assigns a high probability to one wrong label), this allows us to directly measure how much compositional information the model takes into consideration. By effectively punishing models' over-reliance on lexical features, this testing setup could encourage the development of models that are sensitive to compositional clues regarding the pairs' logical relationships.

We show that although our seven models and their variants have comparable performance on standard evaluation (SNLI and MNLI), our new compositionality-sensitivity testing differentiates them by their capability to capture compositional information (e.g., bag-of-words-like models perform worse than sequential models, which in turn perform worse than syntactic-tree based models).
Unlike adversarial evaluation, this setup uses natural examples that are not confined to any specific linguistic context or domain by leveraging existing NLI datasets to the largest extent possible for compositional testing. We hope that this new setup could inspire the collections of datasets that control for lexical-features, to explicitly evaluate the compositional ability of NLI models.

We end by discussing how our compositionality-sensitivity evaluation setup complements other recently proposed evaluation setups.
Specifically, while certain linguistically-driven diagnostic datasets are useful in testing for a model's performance in a specific realistic setting, model-driven evaluations such as ours gives insight into why models succeed and fail in these specific linguistic scenarios.
The main contributions of this paper are three-fold:
1) we introduce two new adversarial setups that expose current state-of-the-art models' inability to process simple sentence-level semantics when lexical features give no information;
2) we rigorously test and expose the limits of standard and adversarial evaluations;
3) we propose a novel compositionality-sensitivity test that analyzes a model's ability to recognize compositional semantics beyond the lexical level, and show its effectiveness in separating models based on architecture.

\label{sec:introduction}

\section{Models and Motivation}

In this section, we start with an overview of the designs of seven recently proposed, high-performing natural language inference models and outline several commonalities that are counter-intuitive. 
We argue that these shared traits hint at their over-reliance on lexical features for prediction, and they are not modeling the compositional nature of language. This motivates our further investigation in later sections.

\subsection{NLI Models}
\label{subsec:nli_approaches}
\begin{table}[t]
\centering
\begin{tabular}{cccc}
\toprule
	\bf Model & \bf SNLI & \bf Type & \bf Representation\\
\midrule
    RSE & 86.47 & Enc & Sequential \\
    G-TLSTM & 85.04 & Enc & Recursive (latent) \\
    DAM & 85.88 & CoAtt & Bag-of-Words \\
    ESIM & 88.17 & CoAtt & Sequential \\
    S-TLSTM & 88.10 & CoAtt & Recursive (syntax) \\
    DIIN & 88.10 & CoAtt & Sequential \\
    DR-BiLSTM & 88.28 & CoAtt & Sequential \\
\bottomrule
\end{tabular}
\caption{Summary of the models we evaluate, including their performance, type, and sentence representation.\\
`Enc' = Sentence Encoder `CoAtt' = Co-Attention Model}
    \label{tab:model_description}
\end{table}

\begin{figure*}[!t]
\centering
\includegraphics[width=0.7\textwidth]{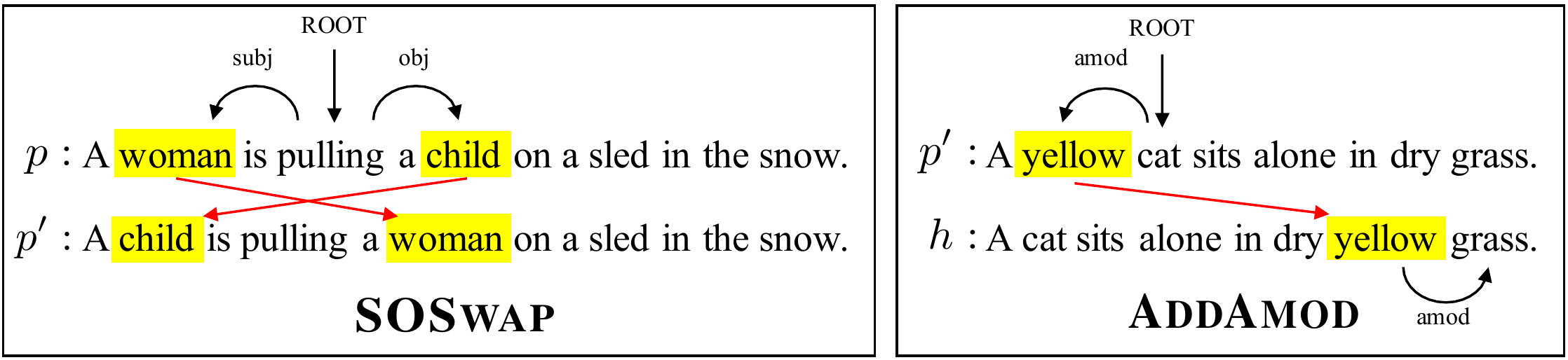}
\caption{Examples of adversaries generated for our experiments. On the left, we have an example of the \subjobjswap\ adversary, where the swapped subject and object are marked in yellow in $p'$. On the right, we have an example of the \amodswap\ adversary, where the added adjective modifier is marked in yellow.}
\label{fig:adversaries_examples}
\end{figure*}

Many different models have been proposed for the NLI task; they all fall under one of two broad categories:
sentence encoding-based (sentence encoders) or co-attention based.
Sentence encoders independently encode each sentence as a fixed-length vector, and then make a prediction, 
while co-attention models make inferences by jointly processing the premise and hypothesis.
The constraint of independent processing for sentence encoders was put in place to encourage the development of effective fixed-length sentence representations generalizable to higher-level tasks.
However, co-attention models with recursive or sequential modeling have achieved much better performance on popular NLI datasets.
In this paper, we analyze 7 different models spanning both categories, which are, or were, state-of-the-art in their respective category.\footnote{{\scriptsize\url{https://nlp.stanford.edu/projects/snli/}\\ \url{https://repeval2017.github.io/shared/}}}  We give a brief description of each model below (see~\tabref{tab:model_description}):
\\
\noindent\textbf{RSE }Residual Sentence Encoder~\cite{RSE} is an encoding-based model that first uses multiple layers of residually-connected BiLSTM to encode the tokens in a sentence and then obtain the sentence's fixed-length representation by max pooling over RNN's hidden states from all timesteps.
It is one of the top performing sentence encoders on the Multi-NLI dataset.
\\
\noindent\textbf{G-TLSTM }
Gumbel-TreeLSTM~\cite{choi2017learning} is a recursive encoding-based model that learns latent-tree representations for sentences via reinforcement learning.
\\
\noindent\textbf{DAM } 
Decomposable Attention Model \cite{decomposable} is a light-weight co-attention model that performs cross-attention at the word level with decomposable matrices.
\\
\noindent\textbf{ESIM } 
Enhanced Sequential Inference Model~\cite{ESIM} is a strong co-attention model that uses BiLSTM to encode tokens within each sentence, and perform cross-attention on these encoded token representations.
\\
\noindent\textbf{S-TLSTM }
Syntactic TreeLSTM~\cite{ESIM} is identical to ESIM except it encodes sentence tokens via a TreeLSTM based on the dependency parse instead of sequential BiLSTM. It is the highest performing NLI model with a recursive component.
\\
\noindent\textbf{DIIN } 
Densely Interactive Inference Network \cite{DIIN} is a novel co-attention model that extracts phrase-level alignment features using densely connected convolutional layers a word-level interactive matrix.
\\
\noindent\textbf{DR-BiLSTM }
Dependent Reading Bidirectional LSTM \cite{DRLSTM} is a model that modifies on ESIM with a dependent reading mechanism that encodes each sentence conditioned on the other.

We were able to obtain original implementations for RSE, G-TLSTM, S-TLSTM and DIIN. We used our own implementation for the other three models and were able to achieve comparable results on standard evaluation sets.

\subsection{Motivation}
Many top-performing sentence encoders (such as RSE) use max-pooling as the final layer to encode the sentences~\cite{repeval2017}, and except DIIN, most top-performing co-attention models calculate cross-alignment on the RNN hidden state of each token.
These design trends are counterintuitive because max-pooling and attention mechanisms are communicative operations which are not affected by word order, making the RNN layers the only way for the models to capture the compositional structure of sentences.
However, past studies have shown that RNNs (especially sequential RNNs) are insufficient for effectively capturing logical compositional structure that is often present in NLI~\cite{can_neural_understand_entail@2018,listops@2018}.
These indicate an over-focus on lexical information in neural NLI approaches which is very different from how humans approach the task.
\label{sec:motivation}

\section{Adversarial Evaluation}
\begin{table}[!t]
\centering
\scalebox{0.88}{
\begin{tabular}{c|c|ccc|ccc}
    & \multicolumn{1}{|c|}{SNLI} & \multicolumn{3}{|c|}{\subjobjswap} & \multicolumn{3}{|c}{\amodswap}\\
    Model & dev & E & \textbf{C} & N & E & C & \textbf{N}\\\hline\hline
    RSE & 86.5  &
        \textbf{92.5} & 2.1 & 5.5 & \textbf{95.2}
          & 0.2 & 4.6 \\
    G-TLSTM & 85.9 & \textbf{97.2}
    	& 1.2 & 1.5 & \textbf{95.9}
        & 1.2 & 2.9 \\
    DAM & 85.0 &
        \textbf{99.7} & 0.3 & 0.0 & \textbf{99.9} & 0.0 & 0.1 \\
    ESIM & 88.2 &
        \textbf{96.4} & 2.1 & 1.5 & \textbf{85.6}
          & 9.6 & 4.8  \\
    S-TLSTM & 88.1 & 
    	\textbf{92.1} & 4.4 & 3.5 & \textbf{90.4}
          & 1.1 & 8.5 \\
    DIIN & 88.1 &
        \textbf{84.9} & 4.5 & 10.6 & \textbf{55.0}
          &  0.4 & 44.6  \\
    DR-BiLSTM & 88.3 &
        \textbf{89.7} & 5.5 & 4.8 & \textbf{82.1} & 8.9 & 9.0
        \\\hline\hline
    Human
    & - & 2 & \textbf{84} & 14 & 10 & 2 & \textbf{88} \\
\end{tabular}}

    \caption{Model performance on SNLI and \% of predictions on the adversarial test sets. E, C, N indicate the classification where E is entailment, C is contradiction and N is neutral. (Note that \subjobjswap\ mostly creates contradictory pairs, while \amodswap\ mostly creates neutral pairs).}
    \label{tab:adv_results}
\end{table}

We test our intuition that the models do not sufficiently capture the compositional nature of sentences by evaluating them on a couple of rule-based adversaries, where we change the semantics of the sentences by perturbing the compositionality without modifying any lexical features.
We found that none of the models were able to successfully use the compositional difference to reason with these examples.

\subsection{Adversarial Examples}
\label{sec:adv_example}

To test our hypothesis that models are over-reliant on word-level information and have limited ability to process compositional structures, we created adversarial test sets composed of pairs of sentences whose logical relations cannot be extracted from lexical information alone.
Specifically, we conduct experiments with the following two types of adversarial data in which we change the semantics of the sentence by only modifying its compositional structure:
\\
\noindent\textbf{\subjobjswap\ Adversaries }
We take a premise from the SNLI dataset, $p$, that contains a subject-verb-object structure, and create the hypothesis $p'$ by swapping the subject and object.
This results in a contradictory pair as the semantic roles of the premise are swapped in the hypothesis. An example is shown on the left side of \figref{fig:adversaries_examples}.
We were able to create 971 examples of this type.
\\
\noindent\textbf{\amodswap\ Adversaries }
In this setup, we take a premise from the SNLI dataset, $p$, that has at least two different noun entities.
We then pick an adjective modifier from the SNLI dataset that has been used to describe both nouns, and create the premise $p'$ by adding the modifier to one of the nouns, and the hypothesis $h$ by adding it to the other.
This results in a neutral pair as the hypothesis contains additional information and is neither implied nor refuted by the premise.
An example of this is shown on the right side of~\figref{fig:adversaries_examples}.
We were able to create 1783 examples of this type.

The intuition behind both of the adversaries described above is that, while the semantic difference resulting from compositional change is obvious for humans, the two input sentences will be almost identical for models that take no compositional information into consideration.\footnote{To create the adversarial data, we used the Stanford Parser \cite{stanford_parser} from CoreNLP 3.8.0 to get the dependency parse of the sentences, on which we apply our strategires.}

\begin{table*}[t]
\centering

\begin{tabular}{lccccccccc}
\toprule
\multirow{2}{*}{\textbf{Model}} & \multicolumn{3}{c}{\textbf{SNLI}} & \multicolumn{3}{c}{\textbf{MNLI Matched}} & \multicolumn{3}{c}{\textbf{MNLI MisMatched}}
\\
\cmidrule(lr){2-4}
\cmidrule(lr){5-7}
\cmidrule(lr){8-10}
     & Original & BoW & WS & Original & BoW & WS & Original & BoW & WS  \\
\midrule
        RSE & 86.47 & 85.02 & -- & 72.80 & 70.02 & -- & 74.00 & 71.10 & --  \\
        ESIM & 88.17 & 82.37 & 86.79 & 76.16 & 68.98 & 73.70 & 76.22 & 69.77 & 74.20 \\
        DR-BiLSTM & 88.28 & 82.81 & 86.90 & 76.90 & 70.11 & 73.27 & 77.49 & 70.70 & 73.25\\
\bottomrule      
\end{tabular}

\caption{The `Original' columns show results for vanilla models on the resp. validation sets.
The `BoW' column show results for BoW-like variants created replacing their RNNs with fully-connected layers.
The `WS' columns show results for models trained with shuffled input sentences.}
\label{tab:shallow_model_results}
\end{table*}

\subsection{Adversarial Evaluation Results}

We trained our 7 models on the SNLI training set and tested them on the adversarial test sets -- the results are shown in~\tabref{tab:adv_results}.
To ensure that the intuitions behind our adversarial generation algorithms were correct, we conducted human evaluation for a sample of 100 examples for each evaluation set.\footnote{The adversaries are intended to use to highlight models' compositional unawareness and motivate further analysis rather than to be a general-purpose evaluation set. Human evaluation results indicates that the the majority of the data are correct.}
On both experiments, despite a majority of the examples being marked as non-entail by our human evaluators, the models classified them overwhelmingly as entailment, indicating the models' inability to recognize or process compositional semantic information\footnote{Note that DIIN's relatively high performance on \amodswap\ is likely due to its convolutional structure successfully capturing the modifier relationship, but we see that it still fails on adversaries with longer-range dependency requirements such as \subjobjswap.}.
The models' poor performance on these adversarial test sets contrasts sharply with their high performance on standard evaluation, raising doubts on the effectiveness and reliability of standard evaluation.
However, adversarial evaluation as done here has its own issues.
We discuss problems with current evaluation further in the next section.

\label{sec:adversarial_evaluation}

\section{Limitations of Existing Evaluations}
\begin{table}[t]
	\centering
	
\begin{tabular}{lcc}
\toprule
   & \subjobjswap & \amodswap \\
  & E/\underline{C}/N & E/C/\underline{N}\\
\midrule
  None & 96.4/\underline{2.1}/1.5 & 85.6/9.6/\underline{4.8}\\
  \subjobjswap & 0.9/\underline{99.1}/0.0 & 66.7/26.9/\underline{6.5} \\
  \amodswap & 73.1/\underline{1.0}/25.9 & 0.3/0.1/\underline{99.6} \\
\bottomrule
\end{tabular}

    \caption{The percentages of predicting E/C/N by ESIM with different types of
    adversarial training, where an underlined number indicates the accuracy on
    the correct label.}
    \label{tab:adv_retrain_results}
\end{table}

In this section, we show that models' performance on standard evaluation does not reflect their compositional understanding capabilities, which we suspect leads to the lack of focus on this type of modeling in the current literature.

\subsection{Regular Evaluation Limitations}
\label{sec:reg_eval}

The gap in model performance between standard evaluation and adversarial evaluation (see \tabref{tab:adv_results}) indicates the limitations of regular evaluation at testing a model's ability to process sentences' compositional structure. More importantly, regular evaluation fails to \emph{separate or differentiate} models that are relying on lexical pattern-matching from those with deeper compositional understanding.
To further illustrate this point, we conduct the following two experiments in which we intentionally force the models to be unaware of compositional information by either removing RNN connections in their architectures or by randomizing word order during training.

\noindent\textbf{RNN Replacement:} We create strong bag-of-words-like models by replacing RNN layers in RSE, ESIM, and DR-BiLSTM with fully-connected layers, and train them on the standard training set.

\noindent\textbf{Word-Shuffled Training:} We train the ESIM and DR-BiLSTM models with the words of the two input sentences shuffled, such that the compositional information is diluted and hard to learn.

The results of these models and their corresponding variants on SNLI, MNLI matched, and MNLI mismatched development set are shown in \tabref{tab:shallow_model_results},
where we see that their performance is not too far from that of their original, recurrent counterparts. To be specific, there is roughly 6-7 points drop in accuracy when RNNs connections are removed and only 2-3 points drop when words are shuffled during training. These counter-intuitive findings indicate that even a model which only considers shallow lexical features is able to get a decent result on standard evaluation, despite using a mechanism that is very different from human reasoning.

\subsection{Adversarial Evaluation Limitations}
\label{sec:adv_eval}
Although, rule-based adversaries were able to expose the models' lack of knowledge of compositional semantics, they have their own limitations and do not serve well as a general analysis tool.
Due to the recursive nature of language, there are infinitely many ways for compositional information to affect a sentence's meaning, but each type of rule-based adversary only tests for one specific compositional rule.
Thus, success on one type of adversary only demonstrates knowledge of that single rule, and does not indicate general knowledge of compositionality rules.
The easiest way to see this is via adversarial training and data-augmentation: we trained the ESIM model with data augmentation from either type of adversaries,\footnote{We add 20,000 adversarial examples into training at each epoch. Adv-Training data was created from SNLI training set while Adv-Evaluation set was created from SNLI dev set.} and re-evaluated the retrained models on both \subjobjswap\ and \amodswap.
As shown in \tabref{tab:adv_retrain_results}, 
while adversarial data-augmentation leads to improvement on the same type of adversary, it
does not generalize to other types of adversaries.
In fact, we see that focusing on one type of adversarial performance may lead to over-fitting that particular adversary, and hurt overall robustness. For example, in \tabref{tab:adv_retrain_results}, we see that adversarial training with \subjobjswap\ leads to an increase in incorrect `contradiction' predictions on \amodswap , and adversarial training with \amodswap\ actually leads to a decrease in performance on \subjobjswap\ while incorrectly increasing `neutral' predictions.\footnote{While it is true that we can use data augmentation from both types of adversaries to improve performance on both types of evaluations, we can easily come up with a third, different type of adversary (e.g., swapping the verbs between the main sentence and a clause) that is still difficult for the 2-adversarially trained model.
Enumerating rule-based adversaries to cover all frequently-used compositional structural changes in a language is prohibitively costly, as generating high-quality (natural and grammatical) data following a single rule already takes tons of time and resources.} These results indicate that models' success on an enumerable set of adversarial evaluation is still far from validating its general compositional ability.

Thus, we propose an alternative evaluation strategy that leverages existing data to evaluate a model's general compositional understanding capabilities.
\label{sec:limitations_of_existing_evaluations}

\section{Compositionality-Sensitivity Testing}

In this section, we first formulate the role of compositionality in the context of NLI task, and then propose a compositionality-sensitivity testing setup as an analysis tool to explicit reveal how much compositional information the models take into consideration for inference.

\subsection{Problem Formulation}
\label{sec:formulation}

NLI is a complex task with many variables -- almost all previous approaches model the task as the distribution $p(y\mid x)$ of the logical relation $y$ conditioned on the pair of input sentences $x=(\textsc{P}, \textsc{H})$, where $y \in $ \{entailment, contradiction, neutral\} and $\textsc{P}$, $\textsc{H}$ are the premise and hypothesis, respectively.
This conditional distribution is often parameterized by some neural models and trained end-to-end by maximizing the probability of `ground-truth' label.
For the sake of studying models' insensitivity to compositional information,
we consider a factorization of the two input sentences
as tuples $(S_p, \Pi_{h})$ and $(S_{h}, \Pi_{h})$, where $S_p$ and $S_h$ are the sets of tokens that make up the premise $\textsc{P}$ and hypothesis $\textsc{H}$ as lexical factors, and $\Pi_p$ and $\Pi_h$ are the sets of compositional rules that combine those tokens into meaningful sentences as compositional factors.\footnote{Due to the complexity of language, lexical elements are often intertwined with compositional rules and this factorization of $p$ will make $\Pi_p$ intractable in practice.
However, we isolate compositional factors from lexical factors in order to analysis of model behavior.}
A perfect modeling of NLI that is capable of taking all lexical and compositional information into account is formalized as \eqnref{eq:ideal_model}, whereas a entirely bag-of-words (BoW) model is formalized as \eqnref{eq:bag_of_word_model}.
\begin{align}
p(y\mid x) & = f_{\theta} (S_p,S_h,\Pi_p,\Pi_h) \label{eq:ideal_model}  \\
p(y\mid x) & = g_{\theta} (S_p,S_h) \label{eq:bag_of_word_model} 
\end{align}

The models we discuss are neither perfect models nor entirely BoW models, but rather a combination of both, where they are able to detect and use some lexical features and some semantic rules:
\begin{align}
p(y\mid x) & = \hat{f}_{\theta} (\tilde{S}_p,\tilde{S}_h,\tilde{\Pi}_p,\tilde{\Pi}_h)
\label{eq:current_model}
\end{align}
where $\tilde S_p \subseteq S_p$ and $\tilde S_h \subseteq S_h$ are the sets of lexical features of the sentences that the model is capable of using, and similarly $\tilde \Pi_p \subseteq \Pi_p$  and $\tilde \Pi_h \subseteq \Pi_h$ are sets of compositional rules that the model is capable of using.
The issue we explored in previous sections is that current models are overly relying on $S_p$ and $S_h$, but have limited ability to detect and use $\Pi_p$ and $\Pi_h$.
In other words, 
$\tilde \Pi_p \ll \Pi_p$ and $\tilde \Pi_h \ll \Pi_h$.
For instance, the adversaries we created Sec. \ref{sec:adversarial_evaluation} have sentence pairs which have the same lexical elements but different compositional structures, i.e., $S_p = S_h$ but $\Pi_p \neq \Pi_h$.
To an entirely BoW model (\eqnref{eq:bag_of_word_model}), this looks identical to the scenario where the same sentence is repeated twice. Thus in those cases, inferences necessarily require knowledge of compositional information.
This provides intuition into our new evaluation setup: 
\textit{In order to evaluate models' compositionality-sensitivity, we need to evaluate their performance on data which can not be solved by lexical features alone, i.e., cannot be solved by an entirely BoW model.}
We thus seek to evaluate models on a subset of the standard evaluation set that fits this criterion.

\subsection{Approximating BoW Model}
To obtain such a subset, we must first approximate an entirely BoW model.
Specifically, we use a softmax regression classifier that takes in only lexical features for prediction.
More formally,
\begin{align}
v &= h(x)\\
p(c \mid x) &= \frac{\exp(w_c^\top v)}{\sum_{c^{\prime} \in L}\exp(w_{c^{\prime}}^\top v)}
\label{eq:liner-regression}
\end{align}
where $h$ is a function that maps the raw input pair $x$ to its lexical feature vector $v \in \mathbb{R}^d $, $p(c|x)$ is the probability given to label $c$ by the softmax regression classifier.
The lexical feature vector $v$ is an indicator vector that contains the following lexical features from the input pair:
\begin{itemize}
\item Unigrams appearance within the premise.
\item Unigrams appearance within the hypothesis.
\item Word pairs (cross-unigrams) where one appears in the premise and the other in the hypothesis.
\end{itemize}
For unigram and cross-unigram features, we only pick words that are nouns, verbs, adjectives or adverbs to reduce sparsity.
We train the regression model on both SNLI and MNLI and use it to approximate an entirely BoW model.

\subsection{Lexically-Misleading Score}
Since the softmax regression classifier we used is not an entirely BoW model, i.e., it does not capture and use all aspects of lexical semantics.
Examples that it predicted incorrectly might still be solvable with the correct lexical information.
Thus, to preserve the integrity of our evaluation, we further remove examples that the softmax regression classifier is ambivalent about, and only look at examples where the regression model was confidently wrong, i.e., cases where they were `misled' by lexical features.
We do so because in cases where the regression has insufficient lexical knowledge (e.g., rare/unseen words), it is likely going to give a less confident prediction, whereas in cases where the model was misled, it had the lexical knowledge to make a decision, and hence a wrong prediction indicates the need for compositional knowledge.

\begin{figure*}[!t]
\centering
\includegraphics[width=1\textwidth]{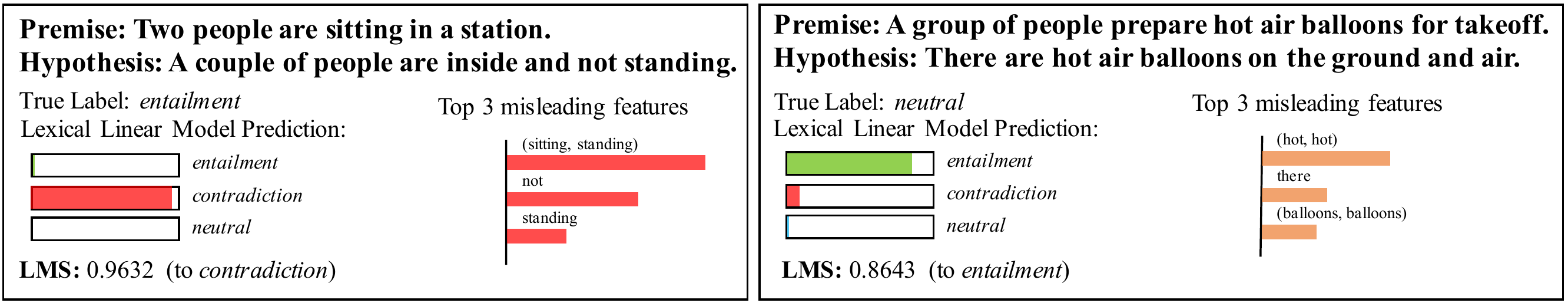}
\caption{Two examples with high LMS. Correct prediction for the 1\textsuperscript{st} example requires recognizing that `not standing' and `sitting' are the same state, rather than focusing on the superficial lexical clues such as `not' and the cross unigram (`sitting', `standing') that both mislead to `contradiction'. For the 2\textsuperscript{nd} example, word-overlap misleads the classifier to predict `entailment'.}
\label{fig:lms_example}
\end{figure*}

Formally, we define the \textbf{Lexically-Misleading Score} (LMS) of an NLI datapoint $(x, c^*)$ as:
\begin{equation}
f_{LMS}(x, c^*) = \max_{c \in L \setminus \{c^*\}} p(c \mid x)
\end{equation}
where $c^*$ is the ground truth label, $p(c\ |\ x)$ is the probability generated by our regression model, and $L = \{\text{entailment, contradiction, neutral}\}$ is the label set.
In other words, $f_{LMS}$ of a data point is the maximum probability the regression gave on an incorrect label. 
The idea behind LMS is that: \textit{the more lexically misleading an example is, the more confident we are that compositional information is required to solve it.}
We thus use LMS to select examples from existing evaluation sets for our evaluation.

\begin{table*}[t]
\centering
\scalebox{0.83}{
\begin{tabular}{rlcccccccccccc}
\toprule
\multirow{3}{*}{\textbf{}} & \multirow{3}{*}{\textbf{Model}} & \multicolumn{4}{c}{\textbf{SNLI}} & \multicolumn{4}{c}{\textbf{MNLI (Matched)}} & \multicolumn{4}{c}{\textbf{MNLI (MisMatched)}}
\\
\cmidrule(lr){3-6}
\cmidrule(lr){7-10}
\cmidrule(lr){11-14}
&  & Whole Dev &  CS\textsubscript{0.5} & CS\textsubscript{0.6} & \textbf{CS\textsubscript{0.7}} &
Whole Dev & CS\textsubscript{0.5} & CS\textsubscript{0.6} & \textbf{CS\textsubscript{0.7}}
& Whole Dev &  CS\textsubscript{0.5} & CS\textsubscript{0.6} & \textbf{CS\textsubscript{0.7}}
\\
\midrule
1 & RSE & 86.47 & 59.01 & 55.59 & 52.73 & 72.80 & 48.48 & 43.57 & 39.62 & 74.00 & 49.30 & 45.84 & 40.85 \\
2 & G-TLSTM & 85.88 & 57.27 & 53.68 & 50.28 & 70.70 & 45.32 & 41.20 & 38.14 & 70.81 & 46.33 & 42.03 & 38.87 \\
3 & \textbf{ESIM} & 88.17 & 62.76 & 58.58 & 55.28 & 76.16 & 52.76 & 49.96 & 48.31 & 76.22 & 54.06 & 51.26 & 48.32 \\
4 & \textbf{S-TLSTM} & 88.10 & 64.60 & 60.57 & \textbf{57.51} & 76.06 & 53.92 & 51.54 & \textbf{48.90}  & 76.04 & 55.60 & 52.40 & \textbf{50.61} \\
5 & \textbf{DIIN} & 88.08 & 64.28 & 60.57 & \textbf{57.17} & 78.70 & 59.49 & 56.12 & \textbf{54.05} & 78.38 & 59.79 & 57.44 & \textbf{53.66} \\
6 & DR-BiLSTM & 88.28 & 62.92 & 58.50 & 55.28 & 76.90 & 55.26 & 52.72 & 50.07 & 77.49 & 57.39 & 55.37 & 53.04\\

\midrule
7 & Human & 88.32 & 81.87 & 80.40 & 80.76 & 88.45 & 86.00 & 86.03 & 86.45 & 89.30 & 85.53 & 85.35 & 84.45 \\
\midrule
8 & Majority Vote & 33.82 & 42.13 & 42.96 & 43.27 & 35.45 & 36.23 & 35.04 & 35.20 & 35.22 & 34.22 & 35.39 & 34.00 \\

\midrule
\multicolumn{13}{c}{Models in which compositional information removed or diluted}\\
\midrule

9 & RSE (BoW) & 85.02 & 52.82 & 47.93 & 43.60 & 70.02 & 40.69 & 34.57 & 31.66 & 71.10 & 43.66 & 38.60 & 34.30 \\
10 & ESIM (BoW) & 82.37 & 48.64 & 44.18 & 40.49 & 68.98 & 38.59 & 33.44 & 30.34 & 69.77 & 41.00 & 35.93 & 32.32 \\
11 & DR-BiLSTM (BoW) & 82.81 & 48.97 & 44.33 & 41.38 & 70.11 & 37.97 & 33.07 & 28.42 & 70.70 & 40.73 & 35.09 & 30.79 \\
12 & ESIM (WS) & 86.79 & 58.41 & 50.61 & 45.49 & 73.70 & 44.20 & 41.20 & 41.09 & 74.20 & 49.39 & 45.39 & 41.77 \\
13 & DR-BiLSTM (WS) & 86.90 & 58.46 & 50.39 & 44.77 & 73.27 & 45.77 & 41.20 & 37.85 & 73.25 & 46.33 & 42.03 & 38.26\\

\bottomrule
    \end{tabular}
}    
\caption{Results of models, human, and majority-vote baseline on different levels of compositionality-sensitivity testing. Results of models with limited compositional information are in the bottem on the table.}
\label{tab:new_metric_results}
\end{table*}

\subsection{Subsampling and Testing}

Given a standard evaluation set and associated `ground-truth' labels, $D = \{(x_i, c_i)\}_{i=1}^N$, we create $\text{CS}_\lambda$, the compositionality-sensitivity evaluation set of confidence $\lambda$:
\[\text{CS}_\lambda = \{(x_i, c_i) \in D \mid f_{LMS}(x_i, c_i) \geq \lambda\}\]
The choice of $\lambda$ represents a trade-off between being confident about the individual examples' ability to test compositionality-sensitivity and keeping a decent sample size of evaluation data.
$\text{CS}_0$ is equivalent to testing on the entire evaluation dataset, whereas $\text{CS}_{0.95}$ (in a 3-way classifier) gives us an extremely small evaluation set (e.g. $\text{CS}_{0.95}$ on SNLI only has 148 examples) with highly misleading lexical features.
Empirically, we found that for SNLI and MNLI, $\lambda = 0.7$ gives a good balance between size of the evaluation set and its ability to test compositionality-sensitivity (e.g., $\text{CS}_{0.7}$ on SNLI has 999 examples).
\figref{fig:lms_example} shows examples of sentence pairs with high LMS from the SNLI validation set that were in $\text{CS}_{0.7}$ for SNLI.\footnote{We release the LMS values of the SNLI and MNLI development set at \scriptsize\url{https://github.com/easonnie/analyze-compositionality-sensitivity-NLI}.}

\subsection{Usage and Limitations}
It is worth noting that we do not wish this subset to be used as a benchmark for models to compete on, but rather an analysis tool to explicitly reveal models' compositionality-awareness. Even though the testing setup has its own limitations such as data sparsity and noisiness, it still serves as an initial step to highlight the problem of compositionality understanding (and gain some important insights into models' behaviors, as shown below), which has been largely unexplored in the current neural literature. But more importantly, we hope that this inspires future works on data collection that explicitly address the issue by adding compositionality requirements and lexical-feature balancing into the collection process.

\subsection{Evaluation of Existing Models on $\text{CS}_\lambda$}
\tabref{tab:new_metric_results} shows the performance of our seven models re-evaluated with $\text{CS}_\lambda$ at different $\lambda$ values.

\noindent\textbf{General Trend:}
We see that in general, model performance decreases as $\lambda$ increases, whereas human performance\footnote{We approximate human performance by the mechanism proposed by~\citet{DIIN}: we choose one of the annotator labels (out of 5) and compare it against the ground truth. Due to noisy data collection procedure, the actual ceiling of human performance should be much higher than this value.} suffered much less with increasing $\lambda$ values.
This is consistent with our hypothesis that there are significant differences between human-style deep reasoning (with both lexical and compositional knowledge) and inference by current models, which overly relies on lexical information. We also noticed that for all the models on SNLI, MNLI matched, and MNLI mismatch dev set, there is a big gap between the accuracy on the whole dev set and those on CS\textsubscript{0.7}. This demonstrates that our models have very limited ability to utilize or even recognize compositional information for semantic understanding. These findings indicate the space and need for further research on structured sentence modeling.
\\
\noindent\textbf{Seqential Model vs. Structured Model}: The results on CS\textsubscript{0.7} differentiates models based on their architectures. More importantly, it explicitly reveals models' compositional understanding which is otherwise largely hidden on the standard evaluation. Specifically, the results for ESIM and S-TLSTM (row 3 and 4) give a clear comparison between sequential and recursive modeling since the two models have the same architecture with the exception that ESIM uses sequential RNN and S-TLSTM uses recursive RNN to encode the sentences. We see that S-TLSTM is better than ESIM on CS\textsubscript{0.7}, despite ESIM getting better results on all three standard evaluation datasets. This indicates that the recursive model with additional syntactic tree input does in fact induce more compostional understanding ability, which is completely invisible if we merely focus on the results of standard evaluation.
Moreover, DIIN (row 5) obtains the best results on all the CS\textsubscript{0.7} subsets, substantially surpassing that of DR-BiLSTM (row 6), the most powerful sequential model in the table. This is also consistent with the intuition that DIIN's convolutional network and phrase-level alignment provide much more compositional information than simple RNN-based sequential models. Another interesting fact is the difference in performance between a recursive model trained with explicit external linguistic supervision (S-TLSTM) and one trained via latent tree learning (G-TLSTM).
We see that S-TLSTM is able to capture compositional information more effectively than G-TLSTM (row 2), which is consistent with findings from diagnostic datasets regarding recursive modeling~\cite{listops@2018}.
\\
\noindent\textbf{Necessity of Compositional Information}: In the lower side of the table (row 9-13), we evaluate models with either severed RNNs connections or word-shuffled training data. The results represent models with limited compositional accessibility or awareness.
As expected, the results on CS\textsubscript{0.7} are similar to or even below the majority vote even though their performance on standard evaluation is on the same level as that of the original models, indicating that compositionality understanding is required to obtain a good result on CS\textsubscript{0.7}.

\label{sec:new_metric}

\section{Related Work and Discussion}
\noindent\textbf{Over-Stability}: 
\citet{adv_squad} used adversarial evaluation to show that models trained on the Stanford QA Dataset \cite{squad} were reliant on syntactic similarity for answering, revealing the over-stability of QA models.
With similar motivation, we study the task of NLI by showing that models are overly focused on lexical features and have limited ability of compositionality.
\\
\noindent\textbf{Existing Analysis on NLI}: 
Previous work on analyzing NLI models \cite{breakingnli_lexical2018,nli_behavior_analysis2018} has focused on models' limited ability in identifying lexical semantics that were rare or unseen during training.
Our work complements theirs by demonstrating models' limited understanding of compositionality encoded in the sentences. \citet{annotating_artifacts}, \citet{poliak2018hypothesis} and \citet{tsuchiya2018performance} concurrently showed hypothesis bias in NLI and RTE datasets. In particular, \citet{annotating_artifacts} also proposed to evaluate models on a harder and better subset of standard evaluation.
We instead focus on exposing models' compositionality-insensitivity by selecting our evaluation dataset based on LMS (lexically-misleading score).
\\
\noindent\textbf{Compositionality}: 
\citet{listops@2018} introduce a dataset to study the ability of latent tree models. \citet{can_neural_understand_entail@2018} introduce a dataset of logical entailments for measuring models' ability to capture the structure of logical expressions against an entailment prediction task. 
\citet{sb_compositionality_evaluating@2018} study the inference behavior of models using sentence embedding on a compositional comparisons dataset. However, we conduct a rigorous study on compositionality-sensitivity, covering a broader range of NLI models and show how to use a filtered subset of existing NLI datasets to test models' compositional ability.
\\
\noindent\textbf{Linguistic Diagnostic Evaluation}: %
Multiple linguistic diagnostic datasets have been published to test NLI models' ability to process certain linguistic phenomena such as coreference, double negation, etc.~\cite{williams2017broad,repeval2017,towards_evaluateNLI@2018,glue@2018}.
These datasets are helpful in that they explore the potential usefulness of existing models by demonstrating their abilities in specific scenarios.
However, the way models approach language might not have any linguistic grounding.
Consider an example where the premise is `We can't not go to sleep.' and hypothesis is `We have to go to sleep.' 
Understanding the first sentence should need compositionality for processing the double negation. However, given that the example's  LMS score is only 0.2376 (which means our BoW regression model was able to solve this correctly), models can solve this particular example via a lexical feature shortcut. Thus, a model resolving of a specific linguistic problem does not necessarily indicate its understanding of the linguistic rule and its generalizability to other compositionality rules. Thus, our work complements linguistic diagnostic datasets well, since a model performing well on both our evaluation and the linguistic diagnostic setup is likely using compositional rules (i.e., human-like reasoning) rather than other pattern-matching procedures to obtain seemingly-compositional behavior.

\label{sec:related_work}

\section{Conclusion}
In this paper, we show that current NLI models achieve misleadingly high results on standard evaluation due to its inability to test models' compositional semantic understanding.
We further show that typical adversarial evaluation is also limited in terms of evaluating generalizability.
Therefore, to encourage the design of models with general understanding capabilities, we propose our compositionality-sensitivity testing that evaluates using compositional information which is not confined to any specific type.
Our work complements other recent advancements in evaluation in the community and we hope that this not only encourages the development of structured compositional-aware models, but also highlights the need of more lexical-feature-controlled data collections for semantically demanding tasks (e.g. NLI), such that they will require not only distributional semantics, which is often captures via large scale unsupervised learning, but also compositional semantics, which tend to be overlooked but a harder problem in the community.
\label{sec:conclusion}

\section*{Acknowledgments}
We thank the reviewers for their helpful comments.
This work was supported by faculty research awards from Verisk, Google, and Facebook.

\bibliography{ref}
\bibliographystyle{aaai}

\end{document}